\documentclass[letterpaper]{article} 
\usepackage{aaai25}  
\usepackage{times}  
\usepackage{helvet}  
\usepackage{courier}  
\usepackage[hyphens]{url}  
\usepackage{graphicx} 
\usepackage{natbib}  
\usepackage{caption} 
\usepackage{algorithm}

\usepackage{newfloat}
\usepackage{listings}
\DeclareCaptionStyle{ruled}{labelfont=normalfont,labelsep=colon,strut=off} 
\floatstyle{ruled}
\newfloat{listing}{tb}{lst}{}
\floatname{listing}{Listing}
\title{Multi-Modal Grounded Planning and Efficient Replanning\\ For Learning Embodied Agents with A Few Examples}
\author{
    Taewoong Kim,
    Byeonghwi Kim,
    Jonghyun Choi$^*$\thanks{\hspace{-2em}$^*$JC is with ECE, ASRI \& IPAI in SNU and a corresponding author.} 
}
\affiliations{
    Seoul National University\\
    \{twoongg.kim, byeonghwikim, jonghyunchoi\}@snu.ac.kr
}

\usepackage{comment}

\usepackage{arydshln}
\usepackage{booktabs}
\usepackage{enumitem}
\usepackage{combelow}
\usepackage{tabularx}
\usepackage{cite}

\usepackage[stable]{footmisc}

\usepackage{multirow}
\usepackage{makecell}
\usepackage{mathtools}
\usepackage{graphicx}
\usepackage{eqparbox,array}

\usepackage{xcolor,colortbl}

\usepackage{subcaption}
\usepackage{sidecap}

\usepackage{nicematrix}

\usepackage{mwe} 
\usepackage{calc} 
\usepackage{pifont} 
\newcommand{\cmark}{\color{blue}{\ding{51}}}
\newcommand{\xmark}{\color{red}{\ding{55}}}

\DeclareMathOperator*{\argmax}{arg\,max}

\usepackage{algpseudocode}
\usepackage{algorithm}

\newcommand{\mcc}[1]{\multicolumn{#1}{c}}

\definecolor{Gray}{gray}{0.90}
\newcolumntype{a}{>{\columncolor{Gray}}r}
\newcolumntype{b}{>{\columncolor{Gray}}c}

\definecolor{azure}{rgb}{0.0, 0.5, 1.0}

\newcommand{\progressone}{\raisebox{-1pt}{\includegraphics[height=9px]{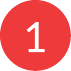}}}
\newcommand{\progresstwo}{\raisebox{-1pt}{\includegraphics[height=9px]{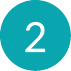}}}
\newcommand{\progressthree}{\raisebox{-1pt}{\includegraphics[height=9px]{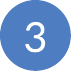}}}

\newcommand{\method}{\mbox{\textsc{FLARE}}\xspace}
\newcommand{\methodfull}{{\textsc{Few-shot Language with environmental Adaptive Replanning Embodied agent}}\xspace}

\RequirePackage{xspace}
\makeatletter
\DeclareRobustCommand\onedot{\futurelet\@let@token\@onedot}
\def\@onedot{\ifx\@let@token.\else.\null\fi\xspace}

\def\eg{\emph{e.g}\onedot} 
\def\Eg{\emph{E.g}\onedot}
\def\ie{\emph{i.e}\onedot}

\def\etc{\emph{etc}\onedot} 
\def\vs{\emph{vs}\onedot}

\renewcommand{\@fnsymbol}[1]{}  
\makeatother

\begin{document}

\maketitle

\begin{abstract}
    Learning a perception and reasoning module for robotic assistants to plan steps to perform complex tasks based on natural language instructions often requires large free-form language annotations, especially for short high-level instructions.
    To reduce the cost of annotation, large language models (LLMs) are used as a planner with few data.
    However, when elaborating the steps, even the state-of-the-art planner that uses LLMs mostly relies on linguistic common sense, often neglecting the status of the environment at command reception, resulting in inappropriate plans.
    To generate plans grounded in the environment, we propose \method (\methodfull), which improves task planning using both language command and environmental perception.
    As language instructions often contain ambiguities or incorrect expressions, we additionally propose to correct the mistakes using visual cues from the agent.
    The proposed scheme allows us to use a few language pairs thanks to the visual cues and outperforms state-of-the-art approaches.
    Our code is available at \url{https://github.com/snumprlab/flare}. 
\end{abstract}

\section{Introduction}

By the rapid advancement in the fields of computer vision, natural language processing, and embodied AI, we are witnessing a significant improvement in key functionalities of robotic assistants that can perform daily tasks.
These functions include navigation~\cite{anderson2018vision, chaplot2017gated, uppal2024spin},
object manipulation~\cite{zhu2017visual, ryu2023diffusion},
and responsive reasoning~\cite{embodiedqa, gordon2018iqa, majumdar2024openeqa}
in simulated 3D spaces~\cite{ge2024behavior, chang2017matterport3d, xia2018gibson, kim2024realfred}.
Practical robotic assistants require all of the aforementioned capabilities to understand language instructions and actively perceive the environment.

\begin{figure}[t!]
    \centering
    \includegraphics[width=\linewidth]{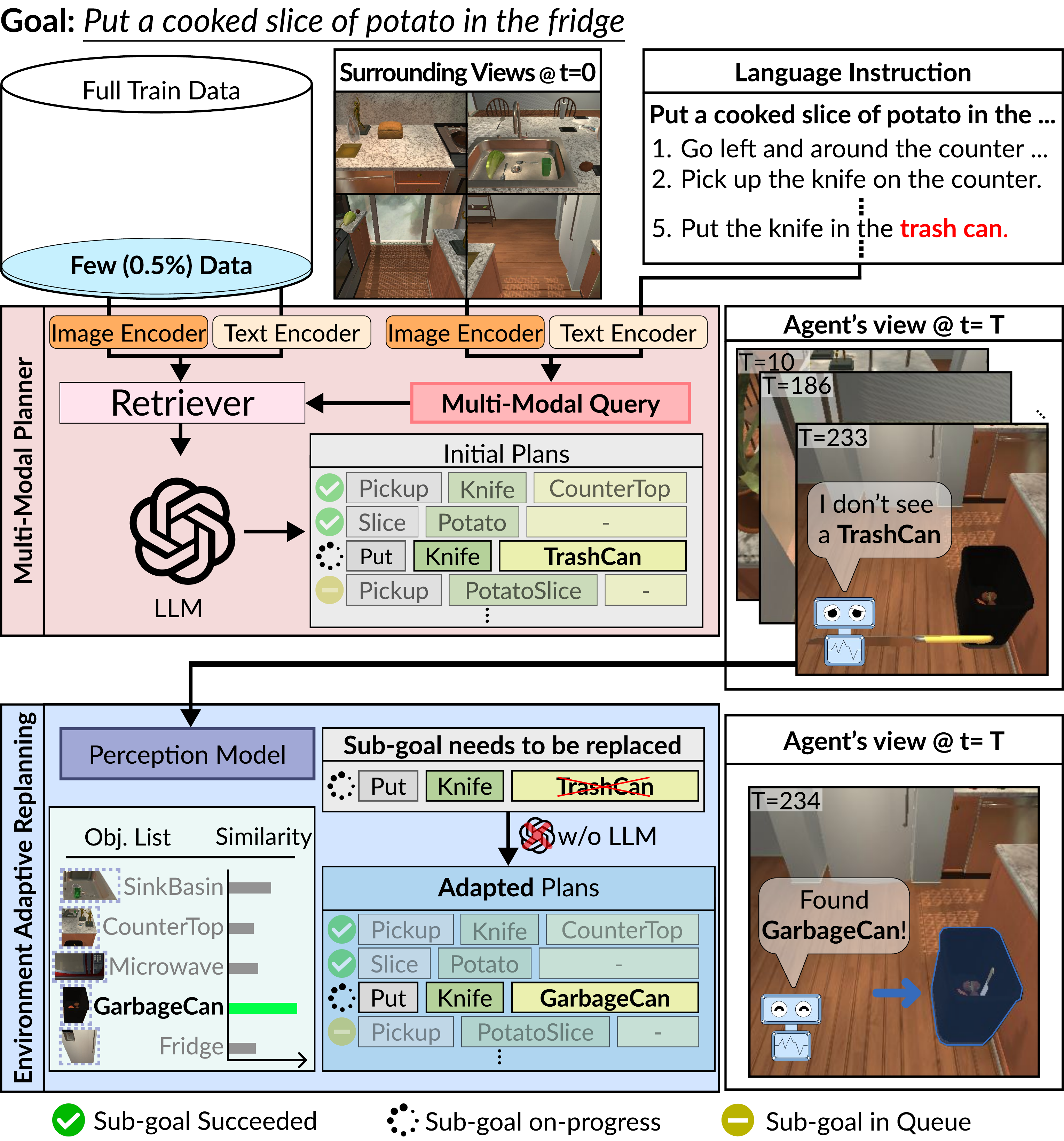}
    \caption{
        \textbf{Overview of the proposed \method.}
        Our agent consists of (1) `Multi-Modal Planner (MMP)' and (2) `Environment Adaptive Replanning (EAR)'.
        MMP takes into account both the agent's initial surrounding views and received instructions to generate a sequence of subgoals by prompting an LLM (\eg, GPT-4).
        When the agent gets stuck while executing a plan, EAR adjusts the ungrounded plan to a physically grounded one with visual cues.
    }
    \vspace{-1.2em}
    \label{fig:Teaser}
\end{figure}

To learn an agent that performs such complex tasks, a straightforward approach is to train an agent in a supervised manner by a large amount of natural language instruction and action pairs~\cite{shridhar2020alfred, ehsani2024spoc, pashevich2021episodic, blukis2021persistent, kim2023context}.
However, annotating instructions and providing expert action sequences (\ie, navigating trajectories) is costly and time-consuming, and thus collecting a sufficient amount of language annotations is often prohibitive.
When data are insufficient, the above-mentioned data-driven approaches would not be effective~\cite{min2021film,inoue2022prompter,suvaansh2023multi, song2022one}.

\begin{figure*}[t!]
    \centering
    \includegraphics[width=0.98\linewidth]{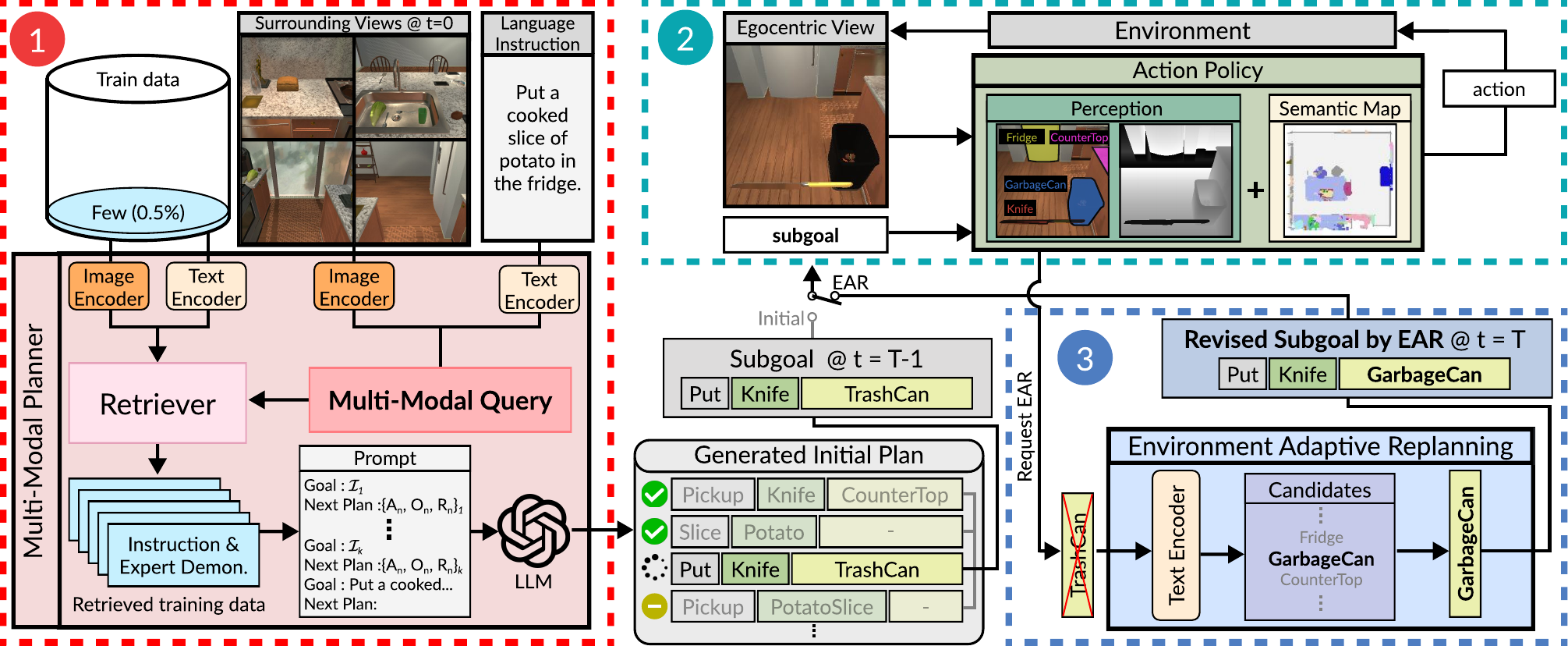}
    \caption{
        \textbf{Detailed architecture of \method.}
        It comprises `Multi-Modal Planner (MMP)' and `Environment Adaptive Replanning (EAR)'.
        \progressone{} MMP retrieves the top $k$ relevant training data pairs with instruction and expert demonstration (indicated with Expert Demon.), based on the agent's initial panoramic surrounding views and language instructions, then plans a sequence of actions through LLMs (\eg, GPT-4) with these examples.
        \progresstwo{} When agent fails to locate the target object (\eg, `TrashCan'), it requests replanning via EAR.
        \progressthree{} Using visual observations and semantic similarity, EAR identifies the most similar object available within the scene and replaces the missing one (\eg, `GarbageCan').
    }
    \vspace{-0.7em}
    \label{fig:architecture}
\end{figure*}

To learn an agent performing a long-horizon task by a small amount of annotated data, recent approaches~\cite{min2021film,inoue2022prompter} exploit manually designed action sequences for specific task types. 
However, such defined actions do not scale to various types of tasks.
Another line of research uses large language models (LLMs) to address insufficient data, capitalizing on the remarkable advances achieved by prior knowledge encoded within LLMs in various domains~\cite{zeng2022socratic, singh2023progprompt, driess2023palme, song2023llmplanner, sarch2023helper, wu2023tidybot}.
In particular, \cite{song2023llmplanner} proposes using LLMs as a high-level planner with a dynamic grounded replanning on top of an existing agent~\cite{blukis2021persistent}, where human-annotated language is very scarce.
But they neglect an environment state that could lead to the generation of implausible plans, since planning often needs to consider the state of the environment when it receives an instruction (\eg, where the agent was located, what is visible from its view, \etc).
To modify the inappropriate plan, \cite{song2023llmplanner} invokes an LLM multiple times with the prompt that includes a list of observed objects to generate a grounded plan.
However, the heavy reliance on large models makes this approach more costly than necessary, as it revises entire sequences when only partial subgoals are incorrect.

To address these issues, we propose \method (\methodfull) which considers the multimodal environmental context (\ie, visual input, and language directive) when it begins to plan the subgoal sequences to complete the household task and efficiently (\ie, partially) correct the plan without using the LLM by using visual input.
To empirically validate the effectiveness of our approach, we adopt a widely used benchmark for embodied instruction following~\cite{shridhar2020alfred}.
We observe that \method can generate plausible plans with only a few, \eg, $100$ language and demonstration pairs, outperforming state-of-the-art methods in our empirical validation by noticeable margins up to $+24.46\%$ absolute gain in the test unseen split.

\noindent We summarize our contributions as follows:
\begin{itemize}
    \item Proposing a multi-modal planner that considers both the environmental status and the language instruction for a long-horizon tasks with a few data.
    \item Proposing a computationally efficient environment adaptive replanner that revises misleading subgoals by visual cues, enabling the generation of plans grounded to the states in the environment without LLM.
    \item Achieving state-of-the-art performance in few-shot settings in the ALFRED benchmark~\cite{shridhar2020alfred} in all metrics.
\end{itemize}


\section{Related Work}
We first review the attempts to use an LLM in robotics, especially for task planning.
Then, we discuss recent approaches to tackle complex instruction-following tasks.

\vspace{-0.5em}
\paragraph{Foundation models for task planning.}
With a recent development in large foundation models (\ie, LLMs and VLMs)~\cite{brown2020language, chen2021evaluating, zhang2022opt, liu2023llava}, they are used as a tool for reasoning~\cite{zeng2022socratic, singh2023progprompt, driess2023palme}, planning~\cite{song2023llmplanner, sarch2023helper, yang2024embodied, szot2024large}, and manipulation~\cite{wu2023tidybot, rss2024moka} in robot systems.
Early approaches in robotic planning~\cite{huang2022language} using LLMs plan the subtasks by iterative enhancement of input prompts.
For example, when the agent does not execute a planned action, \cite{Huang2023Inner} uses multiple environmental feedbacks to adjust the initial plan to recover its failure.

Similarly, \cite{ahn2022can} enabled robot planning with skill affordance value functions for planning.
To directly produce actionable robot policies, \cite{singh2023progprompt, liang2023code} structured a programmatic LLM prompt.
Meanwhile, VIMA~\cite{jiang2023vima} and PaLM-E~\cite{driess2023palme} use multimodal prompts to control robots.

For the purpose of expanding to a variety of tasks, \cite{wang2023prompt} reveals that with well-crafted instructions, LLMs can effectively instruct a quadruped robot in locomotion tasks.
Scaling to open-ended environments~\cite{fan2022minedojo}, agents~\cite{wang2023voyager, zheng2024steveeye} uses LLMs to build a continual learning agent.

While these methods make significant progress in robot planning by using LLMs, they depend on multiple interactions with the LLM to refine or adapt the robot's behavior, leading to heavy inference cost or network overhead if they are used as API calls. 
In contrast, our \method does not use LLMs for replanning to improve computational efficiency in adapting agents to the current environment.

\vspace{-0.5em}
\paragraph{Instruction following embodied agents.}
Embodied instruction following task requires an agent to generate a sequence of actions that align with natural language instructions within a given environment.
Many prior arts~\cite{singh2021factorizing, nguyen2021look, pashevich2021episodic} train an agent in an end-to-end manner, directly generating low-level actions from natural language instructions.
Simultaneously, a templated approach has been proposed for planning a long-horizon tasks~\cite{min2021film, inoue2022prompter}.
While it is data efficient, it is limited to solving the \emph{predefined} tasks and does not generalize to novel tasks.

As the hierarchical or modular planning approach has been shown to be effective in instruction following tasks~\cite{min2021film, blukis2021persistent, kim2023context, xu2024disco}, an attempt to take advantage of LLMs as planners occurred.
\cite{song2023llmplanner, sarch2023helper} use LLM as a high-level planner in such tasks by prompting LLM with few-shot in-context examples, which have been shown to be highly effective~\cite{brown2020language}.
Both methods retrieve several prompting examples based on the distances of the embedded language instructions.


\section{Approach}

Generating executable grounded plans is one of the key components in developing a successful embodied AI agent~\cite{murray2022following, inoue2022prompter, kim2023context}.
State-of-the-art methods~\cite{kim2023context, min2021film, blukis2021persistent, pashevich2021episodic} rely heavily on extensive data, implying that they would not be effective in data-scarce learning scenarios.
However, given the high costs of annotating free-form language instructions, it is desirable to develop a more practical approach to learn an agent using small amounts of data.
In addition to efforts to use LLMs as planners~\cite{ahn2022can, huang2022language, Huang2023Inner, sarch2023helper}, \cite{song2023llmplanner} uses them to learn an agent with a few examples.

However, LLMs do not always generate plausible plans without proper prompt, resulting in the generation of non-sensical or impractical subgoals. 
For example, for the task of `Put a cooked potato in the fridge,' the LLM may tell an agent to `wrap the potato with a foil,' where `wrapping' is not supported by the agent.
Although LLMs create executable plans quite successfully, the inherent ambiguity and lexical diversity of open-vocabulary descriptions often make the connection of language-based instructions to the physical world less clear.
To be specific, an agent that has learned a \textit{sofa} would fail to recognize a \textit{couch}.
This may result in plans that are not well grounded in environments, leaving the agent unable to effectively cope with ungrounded plans when faced with real-world scenarios (\eg, endlessly wandering for an object that is not presented in the scene).
To address this issue, we propose \method that improves task planning for AI agents embodied by using visual and text inputs.
Moreover, our approach uses visual observations from the agent, enabling visually adaptive grounded replanning.

Finally, our agent integrates two proposed components, `Multi-Modal Planner' and `Environment Adaptive Replanning.'
Figure~\ref{fig:architecture} illustrates the architecture of our \method.

\subsection{Multi-Modal Planner}
\label{sec:mmp}

\begin{figure}[t!]
    \centering
    \includegraphics[width=0.98\linewidth]{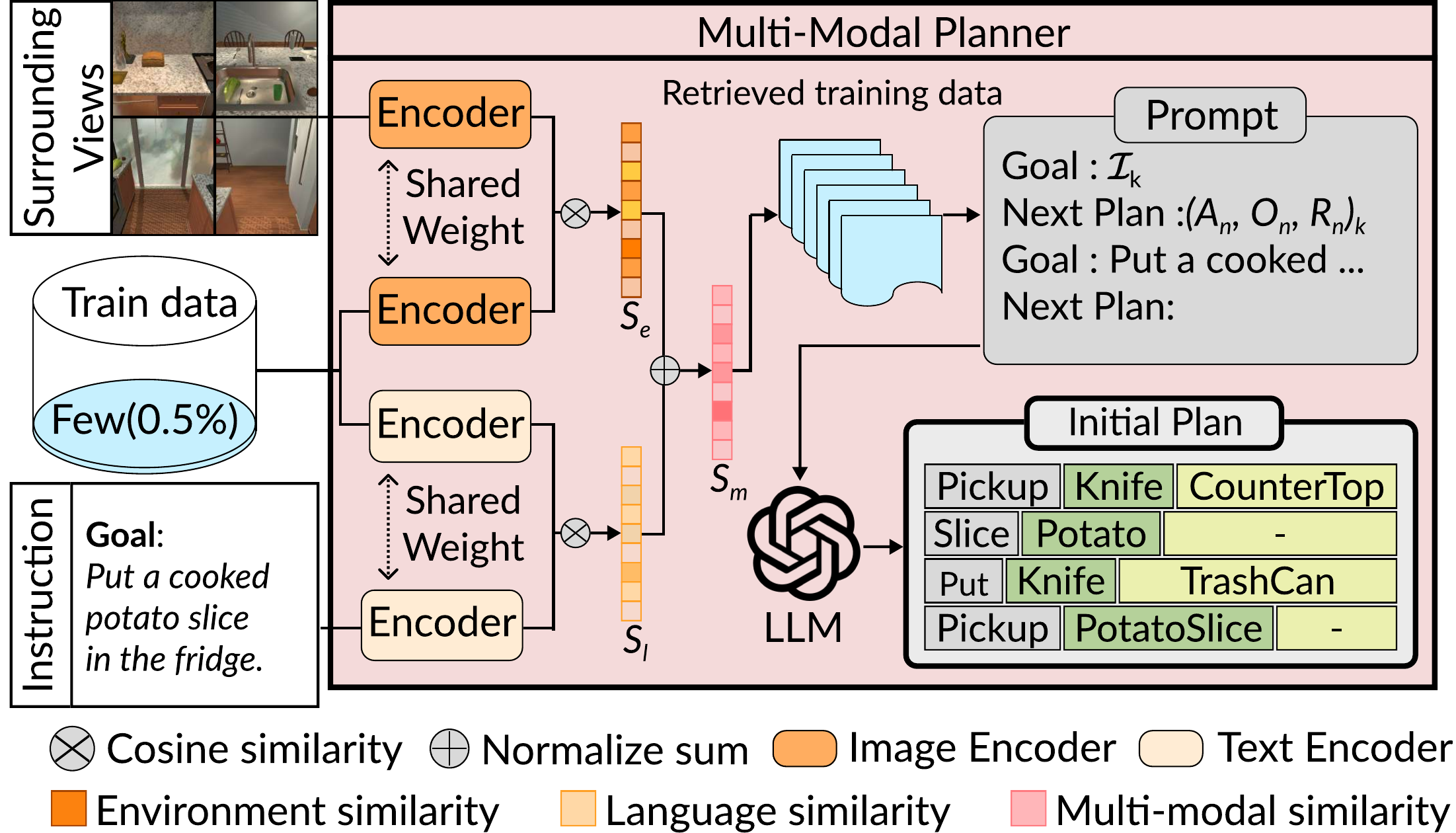}
    \caption{
        \textbf{Multi-Modal Planner.}
        MMP selects top $k$ expert demonstrations based on `multi-modal similarity' (Eq.~(\ref{eq:similarity_multi})) and then converts them into subgoal triplets $(A_n, O_n, R_n)$.
        MMP uses subgoal triplets, along with a text prompt, to guide an LLM in generating task-specific subgoal sequences from natural language instructions.
    }
    \vspace{-0.7em}
    \label{fig:MMP}
\end{figure}

To generate interpretable subgoal sequences for an agent by natural language instructions, LLMs are widely used~\cite{zeng2022socratic, singh2023progprompt, driess2023palme, wu2023tidybot, sarch2023helper}.
For example, \cite{song2023llmplanner, sarch2023helper} retrieves in-context examples from the similarity of language instructions to prompt an LLM.
Inspired by them, we propose `Multi-Modal Planner (MMP)' that considers both the natural language instruction and the agent's egocentric surrounding views at the moment of receiving the command to reflect the environment status only with a few annotated data.
We illustrate an MMP in Figure~\ref{fig:MMP}.

\begin{figure}[t!]
    \centering
    \includegraphics[width=0.98\linewidth]{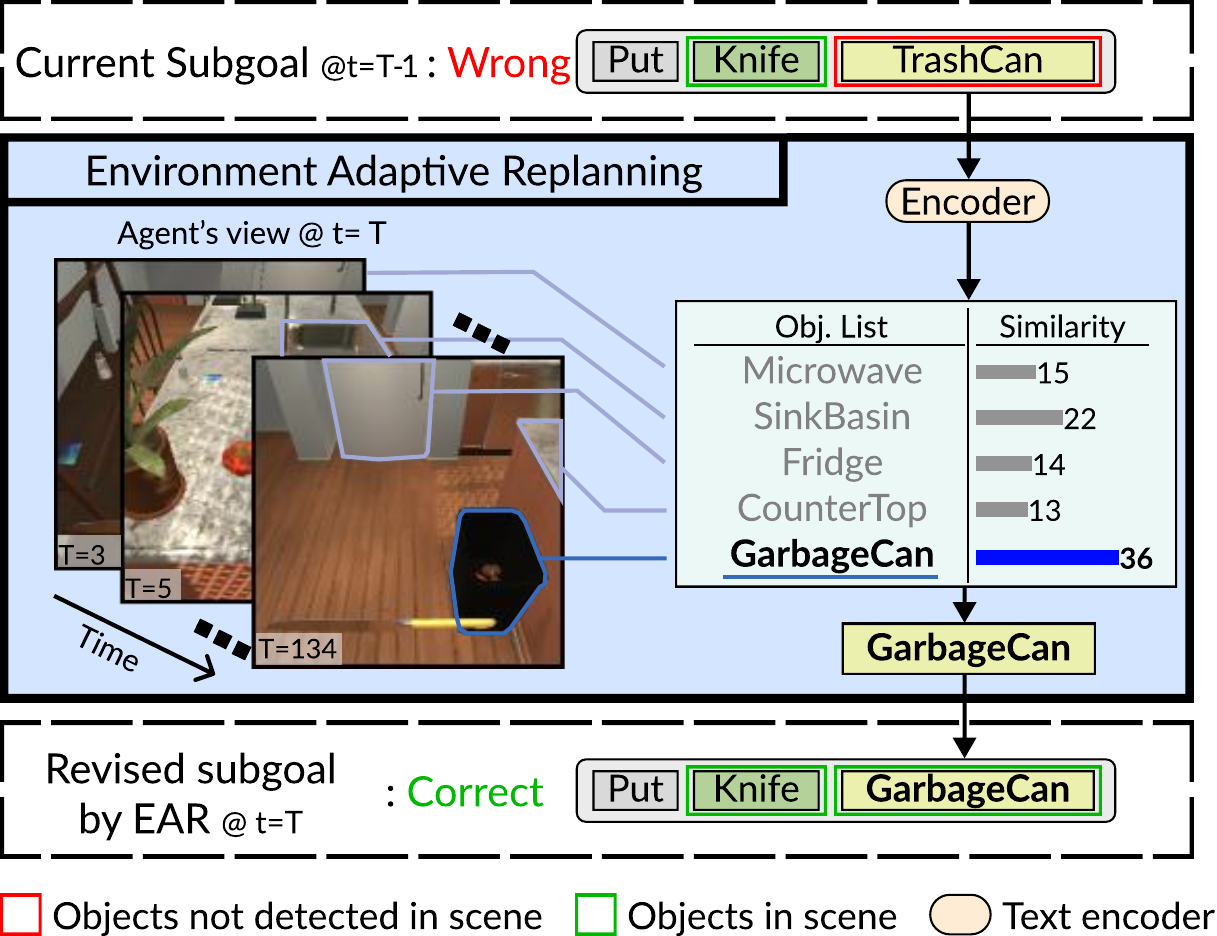}
    \caption{
        \textbf{Environment Adaptive Replanning.}
        EAR corrects a plan by listing detected objects and calculating semantic similarities to replace inaccurately referenced items (\eg, \textit{TrashCan}).
        This ensures that the plan is grounded in the environment.
    }
    \vspace{-0.7em}
    \label{fig:EAR}
\end{figure}

\vspace{-0.5em}
\paragraph{Multi-modal Similarity.}
In-context learning for LLM largely improves model performance for a wide spectrum of language tasks~\cite{brown2020language}. 
It uses explicit context within the prompt to refine the model's comprehension and responsiveness to detailed language instructions.
To capitalize on LLM as a few-shot learner, we need to carefully select examples that are relevant to the task at hand.
When available, such relevant examples assist the LLM's ability to generate appropriate subgoals.
For example, when the task is to `clean a cloth,' it is strategically sound to prompt the model with examples themed in analogy such as `cleaning a fork' or `washing dishes' over unrelated tasks such as `heating apples.'
While \cite{song2023llmplanner, sarch2023helper} achieve this by measuring the distances of the embedded language instructions, they do not consider the environment state (\ie, visual information) when generating a plan.
This can result in inappropriate example selections for the current task (\eg, cutting an apple with a knife when the knife is absent).

To take into account the state of the environment, we use the surrounding views of the agent when receiving a command.
We then embed text instructions with a frozen BERT model~\cite{devlin2018bert} and an image with a frozen CLIP-VIT encoder~\cite{radford2021learning} to gauge how closely each training example aligns with the current task.

Formally, let $ S_l = \{s_{l,1}, s_{l,2}, \ldots, s_{l,N}\} $ and $ S_e = \{s_{e,1}, s_{e,2}, \ldots, s_{e,N}\} $ be language similarities and environment similarities, where $s_{l,i}$ and $s_{e,i}$ represent the language and environment similarity between the current task and the $i^{th}$ example of the training set with the cosine similarities for each embedding vector, respectively. 
Then, we calculate the multi-modal similarities between the current task and the task of the training set ($S_{m}$) by taking the normalized sum of these individual similarity scores as:
\begin{equation}
\label{eq:similarity_multi}
    S_{m} = w_l \cdot \frac{S_l}{\sum_{i=1}^{N} {s_{l,i}}} + w_e \cdot \frac{S_e}{\sum_{i=1}^{N} {s_{e,i}}},
\end{equation}
where $w_l$ and $w_e$ denote the weight of each instruction and environment similarity.
Using the multi-modal similarity score, we retrieve the top $k$ most relevant examples from the training data.
These examples serve as in-context learning examples during the LLM's generation process, thus guiding the LLM to generate a more accurate subgoal.

\vspace{-0.5em}
\paragraph{Subgoal Representation.}
\label{sec:subgoal_triplet}
Translating language instructions into subgoals is one of the key components of robotic reasoning.
\Eg, the task `Move an apple from countertop to a dining table' can be decomposed into subgoals that include both navigation and object interaction, such as [Navigate, CounterTop], [Pickup, Apple], [Navigate, DiningTable], and [Put, DiningTable].
We propose an intermediate subgoal representation using a triplet of [Pickup, Apple, CounterTop] and [Put, Apple, DiningTable] for this representation to reduce the total length of the instruction sets.
Formally, for a given task instruction $\mathcal{I}$, we represent a subgoal as:

\begin{equation}
    S_n = (A_n, O_n, R_n),
\end{equation}
where $n \in \{1,...,K \}$, $K$ is the total number of subgoals in the sequence, $A_n$ denotes high-level actions (\eg, `pick' or `clean'), $O_n$ denotes the target object of the action, and $R_n$ denotes the receptacle where $O_n$ is located.
This approach reduces token usage by $25\%$ compared to~\cite{song2023llmplanner}.

\begin{figure}[t]
\vspace{-1.2em}
\begin{algorithm}[H]
    \small
    \caption{\method algorithm}\label{alg_integration}
    \begin{algorithmic}
        \State \textbf{Input:}
            Time step $t$,
            Subgoal index $k$,
            Uncertainty $u$,
            Uncertainty threshold $\tau$,
            Language instruction $\mathcal{I}$,
            Camera input $\mathcal{C}_t$,
            Subgoal sequences $\mathcal{P}$,
            Semantic map $\mathcal{S}$,
            Detected object set $\mathcal{V}$,
            Current object in interest $O_k$
            
        \State $t, k, u \gets 0$                                                    {\small\color{azure}\Comment{Initialize}}
        \State $\mathcal{P} \gets \text{MMP}(\mathcal{I}, \mathcal{C}_t)$           {\small\color{azure}\Comment{Generate initial plan (Sec.~\ref{sec:mmp})}}
        \State $\mathcal{S} \gets \text{SemanticMapping}(\mathcal{C}_t)$          {\small\color{azure}\Comment{Semantic map (Sec. \ref{sub:semanticmap})}}
        \State $a_{t} \gets \text{ActionPolicy}(\mathcal{P}_k, \mathcal{S})$        {\small\color{azure}\Comment{First action (Sec. \ref{sec:action_policy})}}
        
        \While{$k$ \textless~\text{length}($\mathcal{P}$)}
            \State $\mathcal{C}_t \gets \text{Execute}(a_{t}$)
            \State $\mathcal{S} \gets \text{SemanticMapping}(\mathcal{C}_t)$      {\small\color{azure}\Comment{Update semantic map}}
        
            \State $\mathcal{V}\text{.add}(\text{ObjectDetector}(\mathcal{C}_t))$   {\small\color{azure}\Comment{Update detected object set}}
            \If {$O_k \text{ not in } \mathcal{O}$ }
            \State $u \gets u + 1$
            \If{ $u > \tau$}
            \State $O_k \gets \text{EAR}(O_k, \mathcal{V})$                         {\small\color{azure}\Comment{Replanning (Eq.~(\ref{eq:EAR}))}}
        
            \EndIf
            
            \ElsIf{\text{Complete}($P_k$)}
                \State $k \gets k+1$                                                {\small\color{azure}\Comment{Update subgoal index}}
            \EndIf
            
            \State $t \gets t+1$
            \State $a_{t} \gets \text{ActionPolicy}(\mathcal{P}_k, \mathcal{S})$   {\small\color{azure}\Comment{Next action (Sec. \ref{sec:action_policy})}} 
        \EndWhile
    \end{algorithmic}    
\end{algorithm}
\vspace{-2em}
\end{figure}

\begin{table*}[t!]
    \centering
    \caption{
    \textbf{Comparison with state-of-the-art methods.}
    The path-length-weighteed (PLW) metrics are presented in the parentheses for each metric.
    $^\dagger$We excerpt `SR' and `GC' from \cite{song2023llmplanner}.
    For models that did not report the PLW metric, we noted `N/A' in our comparison.
    }

    \resizebox{.98\linewidth}{!}{
        \begin{tabular}{@{}llaarraarr@{}}
            \toprule
             & & \multicolumn{4}{c}{\textbf{Goal instructions + Sequential instructions}} & \multicolumn{4}{c}{\textbf{Goal instruction only}} \\
             \cmidrule(lr){3-6} \cmidrule(lr){7-10}
            \multirow{2}{*}{\textbf{Setting}} & \multirow{2}{*}{\textbf{Model}}  & \mcc{2}{\textbf{Test Seen}}  & \mcc{2}{\textbf{Test Unseen}} &\mcc{2}{\textbf{Test Seen}}  & \mcc{2}{\textbf{Test Unseen}} \\
                 & & \multicolumn{1}{b}{SR} & \multicolumn{1}{b}{GC} & \multicolumn{1}{c}{SR} & \multicolumn{1}{c}{GC} & \multicolumn{1}{b}{SR} & \multicolumn{1}{b}{GC} & \multicolumn{1}{c}{SR} & \multicolumn{1}{c}{GC} \\

            \cmidrule{1-10}

            \multirow{8}{*}{\hspace{0.3cm}\rotatebox[origin=c]{90}{\parbox[c]{2cm}{\centering\textbf{Few-shot (0.5\%)}}}} & {HLSM~\cite{blukis2021persistent}$^\dagger$}       & $0.82$ ($N/A$)          & $6.88$ ($N/A$)       & $0.61$ ($N/A$)         & $3.72$ ($N/A$) & $N/A$ & $N/A$ & $N/A$ & $N/A$ \\
            & {FILM~\cite{min2021film}$^\dagger$}                & $0.00$ ($N/A$)          & $4.23$ ($N/A$)       & $0.20$ ($N/A$)         & $6.71$ ($N/A$) & $N/A$ & $N/A$ & $N/A$ & $N/A$ \\
            & {CAPEAM~\cite{kim2023context}}           & $0.00$ ($0.00$)       & $3.90$ ($2.29$)    & $0.20$ ($0.00$)      & $6.63$ ($2.36$) & $N/A$ & $N/A$ & $N/A$ & $N/A$ \\

            & {LLM-Planner~\cite{song2023llmplanner}} & $18.20$ ($N/A$)         & $26.77$ ($N/A$)     & $16.42$ ($N/A$)         & $23.37$ ($N/A$) & $15.33$ ($N/A$)     & $24.57$ ($N/A$)  & $13.41$ ($N/A$)    & $22.89$ ($N/A$) \\
            \cmidrule(lr){2-2} \cmidrule(lr){3-6} \cmidrule(lr){7-10}
            & {\method-LLaMA2}            & $16.96$ ($4.60$)  & $24.84$ ($8.09$)  & $17.79$ ($5.62$)  & $27.40$ ($9.46$)  & $12.00$ ($3.01$)  & $20.05$ ($7.22$)  & $13.73$ ($4.27$)  & $21.98$ ($8.46$) \\
            & {\method-Vicuna}                & $20.61$ ($6.28$)  & $29.57$ ($10.17$) & $22.04$ ($7.61$)  & $33.57$ ($12.06$) & $16.37$ ($4.57$)  & $23.68$ ($8.84$)  & $18.05$ ($5.98$)  & $26.75$ ($10.75$) \\
            & {\method-GPT-3.5}                & $32.55$ ($12.17$)     & $42.02$  ($16.94$)  & $31.79$ ($12.21$)    & $43.94$ ($17.44$) & $23.48$ ($8.71$)     & $33.40$ ($14.40$)  & $25.38$ ($9.37$)    & $36.02$ ($15.28$) \\
            & {\method-GPT-4 (Ours)}               & $\textbf{40.05}$ ($\textbf{16.68}$) & $\textbf{48.84}$ ($\textbf{21.31}$) & $\textbf{40.88}$ ($\textbf{18.14}$) & $\textbf{51.72}$ ($\textbf{22.78}$) & $\textbf{31.96}$ ($\textbf{12.93}$) & $\textbf{41.36}$ ($\textbf{18.55}$) & $\textbf{32.57}$ ($\textbf{12.72}$) & $\textbf{43.23}$ ($\textbf{18.40}$) \\

            \cmidrule{1-10}
            
            \multirow{3}{*}{\hspace{0.5cm}\rotatebox[origin=c]{90} {\textbf{Full}} } & {HLSM~\cite{blukis2021persistent}}      & $29.94$ ($8.74$)      & $41.21$ ($14.58$)  & $20.27$ ($5.55$)     & $30.31$ ($9.99$) & $25.11$ ($6.69$)      & $35.79$ ($11.53$)  & $16.29$ ($4.34$)     & $27.24$ ($8.45$) \\
            & {FILM~\cite{min2021film}}               &  $28.83$ ($11.27$)      & $39.55$ ($15.59$)  & $27.80$ ($11.32$)    & $38.52$ ($15.13$)  & $25.77$ ($10.39$)     & $36.15$ ($14.17$)  & $24.46$ ($9.67$)     & $34.75$ ($13.13$) \\
            & {CAPEAM~\cite{kim2023context}}          & $51.79$ ($21.60$)     & $60.50$ ($25.88$)  & $46.11$ ($19.45$)    & $57.33$($24.06$) & $47.36$ ($19.03$)     & $54.38$ ($23.78$)  & $43.69$ ($17.64$)    & $55.66$ ($22.76$) \\

            \bottomrule
        \end{tabular}
    }
    \label{tab:comparison_with_sota}
    \vspace{-1em}
\end{table*}

\subsection{Environment Adaptive Replanning}

Despite the emergent ability of planning by large language models (LLMs), they may generate plans that are not well grounded in environments where agents are deployed.
This issue can be attributed to the lexical variation inherent in natural language instructions.
For example, consider the task, ``Place a tray with a butter knife and slice of the fruit on the table.''
To complete the task, an agent needs to find the \emph{fruit} to be sliced.
However, if the agent has not learned the \emph{fruit} object class for navigation during training, this may lead to navigation failure and possibly, task failure.

To address this issue, the proposed `Environment Adaptive Replanning (EAR)' revises subgoals by replacing an undetected object with the most semantically similar object among those observed so far.
To revise the subgoal, EAR first maintains a list of all detected objects that have been observed so far while completing the task.
For each subgoal, if the agent cannot reach a navigation target (\ie, $O_n$ or $R_n$), EAR infers that the specified object is absent from the environment and replaces it with a semantically analogous one.

To replace a current unavailable object with another one, EAR finds the most \textit{semantically similar} object among the candidates (\ie, objects observed so far).
To measure semantic similarity, we compute the cosine similarity of language representations of two object class names.
Specifically, EAR first obtains the language representations of the names of the current target object and observed objects using a pretrained language model~\cite{devlin2018bert, raffel2020exploring, brown2020language}.
Once obtained, EAR then computes the similarity scores of the observed objects with respect to the current target object.

Formally, we compute the similarity scores and obtain the most semantically similar object as following:
\begin{equation}
    \label{eq:EAR}
        V^* = \argmax_{V_i}~S_C(\text{Enc}(O_k),\text{Enc}(V_i)),
\end{equation}
where $V^*$ is an object that maximizes $S_C(\cdot, \cdot)$, $O_k$ is a current object, and $V_i$ is a $i^{th}$ detected objects so far.
$\text{Enc}(\cdot)$ denotes a language encoder, and $S_C(\cdot, \cdot)$ denotes the cosine similarity of the two embeddings.
Note that $O_k$ can be either $O_n$ or $R_n$.
We illustrate the EAR in detail in Figure~\ref{fig:EAR}.

\subsection{Action Policy} 
\label{sec:action_policy}
For object interaction, the agent first navigates to a target object and reaches it in a close vicinity.
For navigation, a viable approach is to use imitation learning~\cite{shridhar2020alfred,singh2021factorizing,pashevich2021episodic,nguyen2021look}.
However, it requires a large number of training episodes for acceptable performance, but collecting these episodes may not always be available, especially in our case where training data collection is often costly and time-consuming.

To avoid this issue, recent approaches~\cite{inoue2022prompter, kim2023context} incorporate deterministic algorithms (\eg, A* algorithm, FMM~\cite{sethian1996fast}, \etc) obstacle-free path planning, leading to significant performance improvements compared to those learned by imitation learning.
Inspired by recent observations, we adopt the deterministic approach~\cite{sethian1996fast} for effective path planning.


\section{Experiments}
\subsection{Experimental Setup}
We employ four large language models for our \method to validate the compatibility of the proposed methods with different models, incorporating both proprietary and open-source models.
Specifically, we use GPT-4 and GPT-3.5 as proprietary models, and LLaMA2-13B~\cite{touvron2023llama} and Vicuna-13B~\cite{zheng2023judging} as open-source models.
We select $k = 9$ in-context examples, following~\cite{song2023llmplanner} for a fair comparison with it and set $w_l$ and $w_e$ to the same values in equation~(\ref{eq:similarity_multi}) to treat each modality equally.
 
\subsection{Dataset and metrics}
We evaluate the effectiveness of our \method in the ALFRED~\cite{shridhar2020alfred} benchmark.
It requires agents to complete household tasks based on language instructions and egocentric observations within interactive 3D environments~\cite{ai2thor}.
Both validation and test sets include \textit{seen} and \textit{unseen} scenarios, where the \textit{seen} scenario is part of the training data, while the \textit{unseen} scenario represents a new and unfamiliar environment for evaluation.

To evaluate the efficiency of \method where human language pairs are scarce, we followed the same few-shot setting$(0.5\%)$ as in the previous work~\cite{song2023llmplanner}.
For a fair comparison with the previous methods, we use the same number of examples~\cite{song2023llmplanner} (\ie, 100 examples).
The selected 100 examples contain all 7 task types for fair representations of 21,023 training examples.

For evaluation, we follow the same evaluation protocol as \cite{shridhar2020alfred}.
The primary metric is a success rate (SR), measuring the percentage of completed tasks.
A goal-condition success rate (GC) measures the percentage of satisfied goal conditions.
Furthermore, we assess the efficiency of agents penalizing SR and GC (\ie, PLWSR and PLWGC) with the path length of a trajectory taken by the agents.
More details on the dataset and metrics are provided in Section~\ref{sup:benchmark_detail}.

\subsection{Comparison with State of the Arts}
We first compare our method with state-of-the-art methods~\cite{blukis2021persistent, min2021film, kim2023context, song2023llmplanner} and summarize the result in Table~\ref{tab:comparison_with_sota}.
Following~\cite{min2021film, kim2023context, blukis2021persistent}, we report the performance of agents using 1) only a goal statement, denoted by `Goal instruction only,' and 2) both goal statement and step-by-step instructions, denoted by `Goal instructions+Sequential instructions.'

First, we observe significant performance drops from the full-shot setting to the few-shot setting from methods that require a large amount of data to train planners (HLSM, FILM, and CAPEAM).
This implies that learning task-performing agents with limited training examples poses a significant challenge, as this data scarcity can hinder the learning of models with diverse tasks, objects, and environments, implying challenging generalization.

We then compare the results with very recent work using LLMs~\cite{song2023llmplanner} that learns tasks only with a few training examples.
We explore both proprietary and open source language models, including comparative models of lower performance, as shown in~\cite{zheng2023judging}.

Despite with a relatively less capable language models (\ie, LLaMA2~\cite{touvron2023llama}), our proposed agent still outperforms in all metrics in unseen environments in both `Goal instructions + Sequential instructions' and `Goal instruction only,' implying its effectiveness.
Furthermore, using better language models such as GPT-4 can notably improve our agent by large margins up to $24.46\%$ as expected.

\begin{table}[t!]
    \centering
    \caption{
    \textbf{Planner accuracy comparison.}
    `Seen Acc.' and `Unseen Acc.' denote planner accuracies in valid seen and unseen folds.
    A plan is regarded as correct when it is aligned with the ground-truth plan.
    To solely compare planner accuracy, we omit replanning in LLM-Planner and \method, denoted with `Static' and `w/o EAR'.
    }
    \resizebox{1\linewidth}{!}{
    \begin{small}        
        \begin{tabular}{@{}clbc@{}}
            \toprule
            \multirow{1}{*}{LLM} & \multirow{1}{*}{Method}                  & \multicolumn{1}{b}{\textbf{Seen Acc.}}    & \multicolumn{1}{c}{\textbf{Unseen Acc.}} \\
            \midrule

            LLaMA2 & LLM-Planner (Static)~\cite{song2023llmplanner} & 0.006 &  0.002   \\
            LLaMA2 & \textbf{\method (w/o EAR)} & \textbf{18.54} &  \textbf{22.29}\\
            \midrule
            Vicuna & LLM-Planner (Static)~\cite{song2023llmplanner} & 8.17 &  7.06   \\
            Vicuna & \textbf{\method (w/o EAR)} & \textbf{24.51} &  \textbf{33.62}\\
            \midrule
            GPT-3.5 & LLM-Planner (Static)~\cite{song2023llmplanner} & 29.78 &  31.67   \\
            GPT-3.5 & \textbf{\method (w/o EAR)} & \textbf{46.10} &  \textbf{55.66}\\
            \midrule
            GPT-4 & LLM-Planner (Static)~\cite{song2023llmplanner} & 31.54 &  30.12   \\
            GPT-4 & \textbf{\method (w/o EAR)} & \textbf{61.34} &  \textbf{67.48}\\
            
            \bottomrule
        \end{tabular}
    \end{small}    
    }
    \label{tab:planner_acc}
\end{table}

\begin{table}[t!]
    \centering
        \caption{
        \textbf{Ablation study.}
        PLW metrics are presented in the parentheses for each metric.
        MMP and EAR each denotes `Multi-Modal Planner' and `Environment Adaptive Replanning,' respectively.
        We observe that each component contributes to agent's performance.
    }
    \resizebox{1.00\linewidth}{!}{
    \begin{small}
        \begin{tabular}{@{}cccbbcc@{}}
            \toprule
            \multirow{2}{*}{\#}
                & \multirow{2}{*}{MMP} & \multirow{2}{*}{EAR}
                & \mcc{2}{\cellcolor[HTML]{FFFFFF} \textbf{Test Seen}} &  \mcc{2}{\textbf{Test Unseen}} \\
                & & & \multicolumn{1}{b}{SR} & \multicolumn{1}{b}{GC} & \multicolumn{1}{c}{SR} & \multicolumn{1}{c}{GC} \\
            \midrule
            \multicolumn{1}{c}{($a$)} & \cmark & \cmark & $32.55$ ($12.17$) & $42.02$ ($16.94$)     & $31.79$ ($12.21$)    & $43.94$ ($17.44$) \\
            \midrule
            \multicolumn{1}{c}{($b$)} & \xmark & \cmark & $30.20$ ($12.13$) & $41.26$ ($17.27$)     & $30.35$ ($11.62$)    & $42.40$ ($16.66$) \\
            \multicolumn{1}{c}{($c$)} & \cmark & \xmark & $30.79$ ($11.98$) & $40.20$ ($16.51$)     & $30.28$ ($12.01$)    & $42.48$ ($17.03$) \\
            \multicolumn{1}{c}{($d$)} & \xmark & \xmark & $28.05$ ($11.48$) & $38.64$ ($16.23$)     & $28.58$ ($11.82$)    & $39.92$ ($16.13$) \\
            
            \bottomrule
        \end{tabular}
    \end{small}
    }
    \label{tab:ablation_proposed_components}
    \vspace{-0.7em}
\end{table}

\vspace{-0.4em}
\paragraph{Planner Accuracy Comparison.}
To investigate the performance of the initial planner that generates action sequences, denoted by \emph{static} planning, we compare accuracy of our agent and recent LLM-based planning methods~\cite{song2023llmplanner,ahn2022can} by removing their respective replanning strategies and report the result in Table~\ref{tab:planner_acc}.

To isolate LLM's effect in planning, we validate methods using different LLMs.
We observe that our agent equipped with MMP, denoted by `\method (w/o EAR),' consistently outperforms prior work~\cite{song2023llmplanner} by noticeable margins in accuracy for both seen and unseen environments across the LLMs, implying that the improvement of our MMP is not attributed to a particular LLM choice.

\subsection{Ablation Study}
We conduct a quantitative ablation study to analyze components proposed in \method and summarize the result in Table~\ref{tab:ablation_proposed_components}.
We choose GPT-3.5 over GPT-4 as the language model due to the latter's significantly higher token generation cost.

\begin{figure}[t!]
    \centering
    \includegraphics[width=0.94\linewidth]{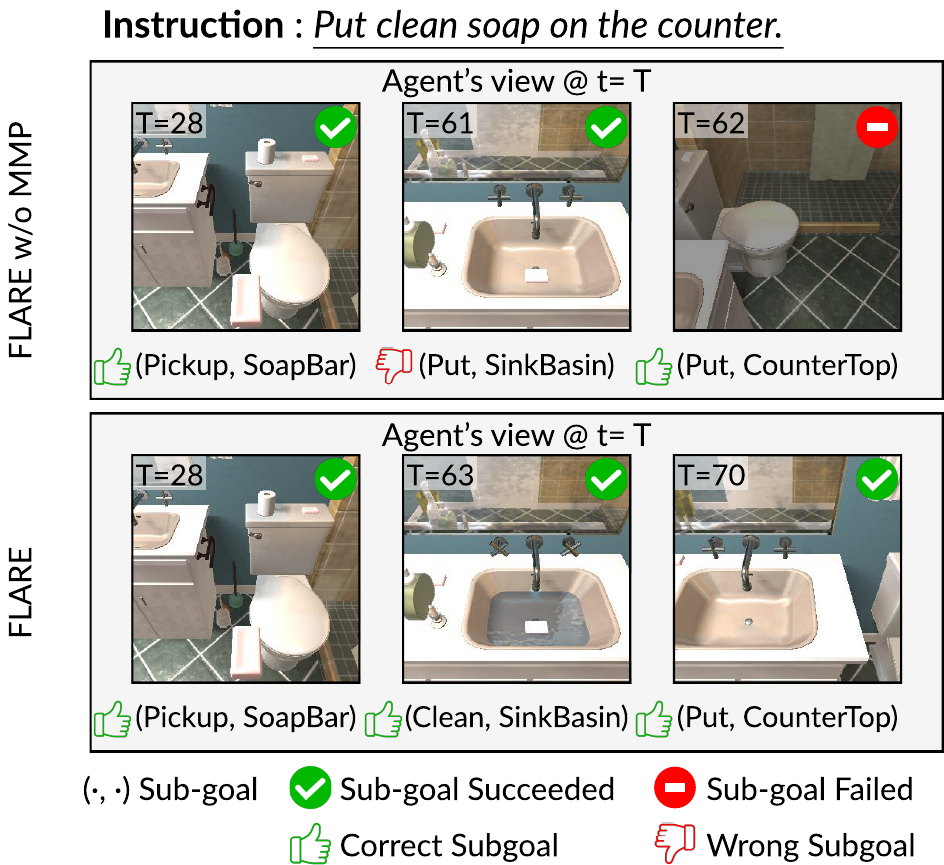}
    \caption{
        \textbf{Benefits of proposed multi-modal planner (MMP).}
        An agent without MMP misinterprets the task, simply placing a \textit{SoapBar} in the \textit{SinkBasin}.
        In contrast, an agent with MMP seems to comprehend an objective of \textit{cleaning}, generating a plausible plan and subsequently completing the task successfully.
    }
    \vspace{-0.8em}
    \label{fig:lang_quali}
\end{figure}

\vspace{-0.5em}
\subsubsection{Without Multi-Modal Planner.}
First, we ablate the `MMP' from our method and the agent considers unimodal similarity to retrieve in-context examples from the dataset, neglecting the environment state for planning.
Without the proposed component, we select in-context examples based on instruction similarity.
Since a prompt reflects a single modality, the agent may omit environmental cues and misinterpret task requirements, leading to performance drops in both seen and unseen splits, as shown in ($\#(a)$ \vs ($\#b$)).

\vspace{-0.5em}
\subsubsection{Without Environment Adaptive Replanning.}
We then ablate `EAR' from our agent.
Without EAR, an agent cannot handle language variation and often misinterprets natural language instruction, leading to an erroneous subgoal.
We observe noticeable performance drops ($1.76 \%p$, $1.51 \%p$ in SR) in both seen and unseen splits, as shown in ($\#(a)$ \vs ($\#c$)).
This implies that LLMs often fail to generate grounded plans in the environment where the agent is deployed, causing the agent to wander in search of an object that may not be present, eventually leading to task failure.

\vspace{-0.5em}
\subsubsection{Without both.}
Without any of the proposed components, the agent adheres to the initial plan, which may not correspond to the current task.
As expected, our agent without both `MMP' and `EAR' achieves the lowest performance among the agents equipped with either or both ($\#(d)$ \vs ($\#a,b,c$)).
Furthermore, we observe that using both multi-modal planning and adaptive replanning of the environment improves performance compared to using only either of them ($(\#(d)$ $\rightarrow$ $\#(b, c)$) \vs $(\#(d)$ $\rightarrow$ $\#(a)$), implying that both components are complementary to each other.

\begin{figure}[t!]
    \centering
    \includegraphics[width=0.98\linewidth]{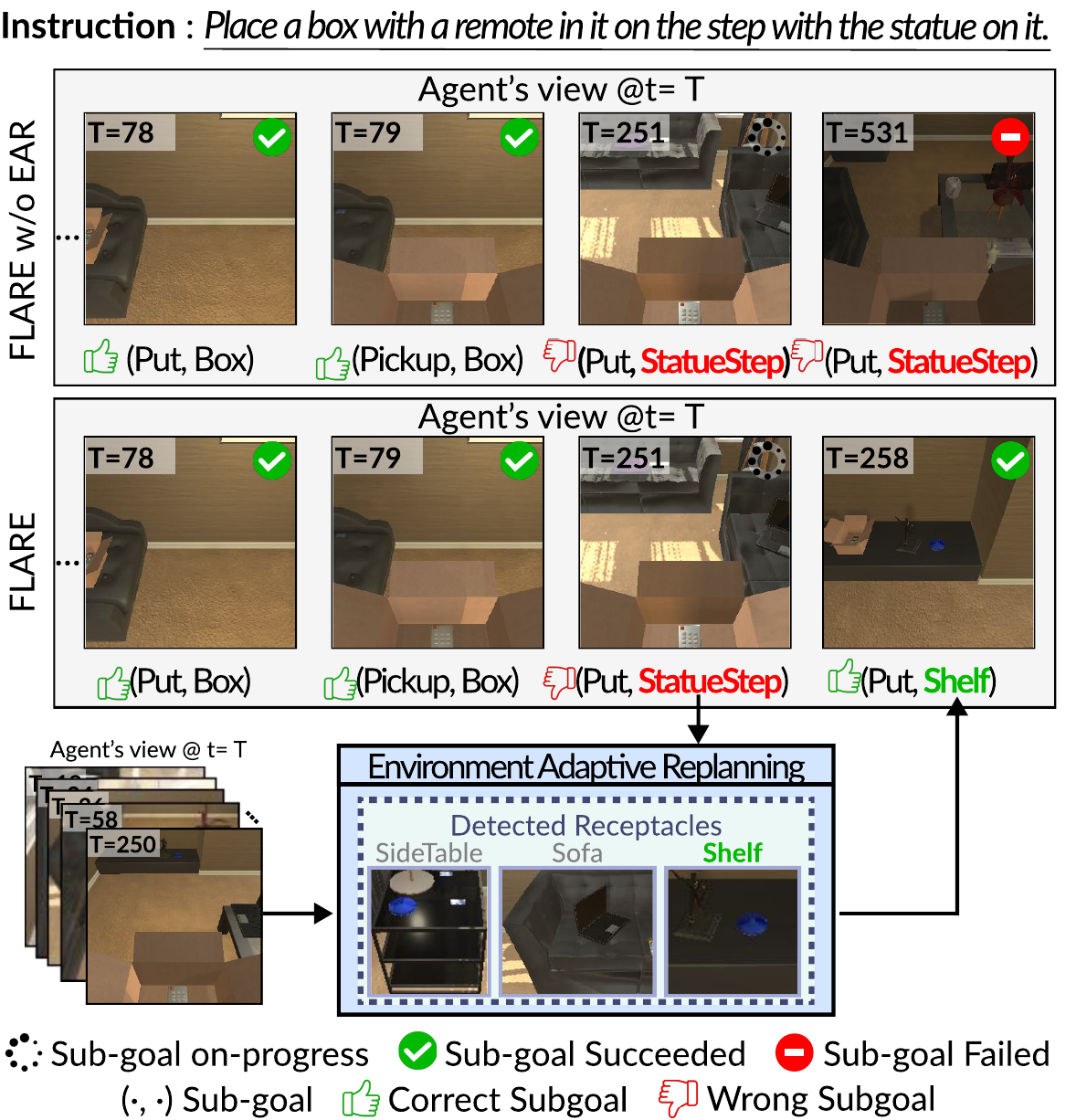}
    \caption{
        \textbf{Benefits of proposed environment adaptive replanning (EAR).}        
        If an agent fails to find the specified object for a certain step, the EAR adapts plan.
        The EAR measures a similarity between misleading and detected objects during navigation.
        It then selects the most semantically similar object to replace the previous one.
    }
    \vspace{-0.7em}
    \label{fig:corr_quali}
\end{figure}

\subsection{Qualitative Analysis}
We analyze our method with several qualitative results and illustrate the result in Figure~\ref{fig:lang_quali},~\ref{fig:corr_quali} and in Section~\ref{sup:add_quali}.

\vspace{-0.5em}
\subsubsection{Multi-Modal Planner.}
To investigate the advantage of multi-modal planning, we present a qualitative example in Figure~\ref{fig:lang_quali}.
As MMP retrieves relevant examples for the current task with multi-modal queries, it would encourage a large language model to generate more plausible subgoal.

We observe that the agent without MMP generates inappropriate subgoals, failing to understand the task context of the language instructions, which leads to placing a \textit{SoapBar} on the \textit{SinkBasin}. 
Subsequently, the agent fails to proceed with the generated subgoal sequence, as it cannot execute the \emph{Put} action with an empty hand.
In contrast, the agent equipped with MMP appears to succeed in extracting prior knowledge from the LLM.
The agent generates satisfying subgoal sequences to pick up a \textit{SoapBar}, clean it in a \textit{SinkBasin}, and finally place it on the \textit{CounterTop}.

\vspace{-0.5em}
\subsubsection{Environment Adaptive Replanning.}
We then investigate the benefit of EAR.
It adapts an unrounded plan with visual cues when the agent fails to locate the specified object.

Due to the various ways in referring to the object, an LLM confused by such diversity may create ungrounded subgoals.
For example, Figure~\ref{fig:corr_quali} shows a scenario where the agent is asked to place a \textit{Box} on `\textit{the step with the statue on it}.'
An LLM that maximizes the given information generates \textit{(Put, Box, StatueStep)} as a subgoal, causing the agent to wander around looking for a \textit{StatueStep} while holding a \textit{Box}.

We observe that the agent without EAR could not distinguish whether the current subgoal is inappropriate (\ie, \textit{StatueStep} does not exist), as it endlessly wanders around the scene and fails to specify receptacle object.
On the contrary, an agent with EAR starts to search for a \textit{StatueStep} initially and notices that \textit{StatueStep} may not be present in the scene.
After replacing the inappropriate object with the most relevant objects presented in the scene (\ie, \textit{StatueShelf} $\rightarrow$ \textit{Shelf}), agent now executes the revised subgoal.

\begin{figure}[t!]
    \centering
    \includegraphics[width=0.94\linewidth]{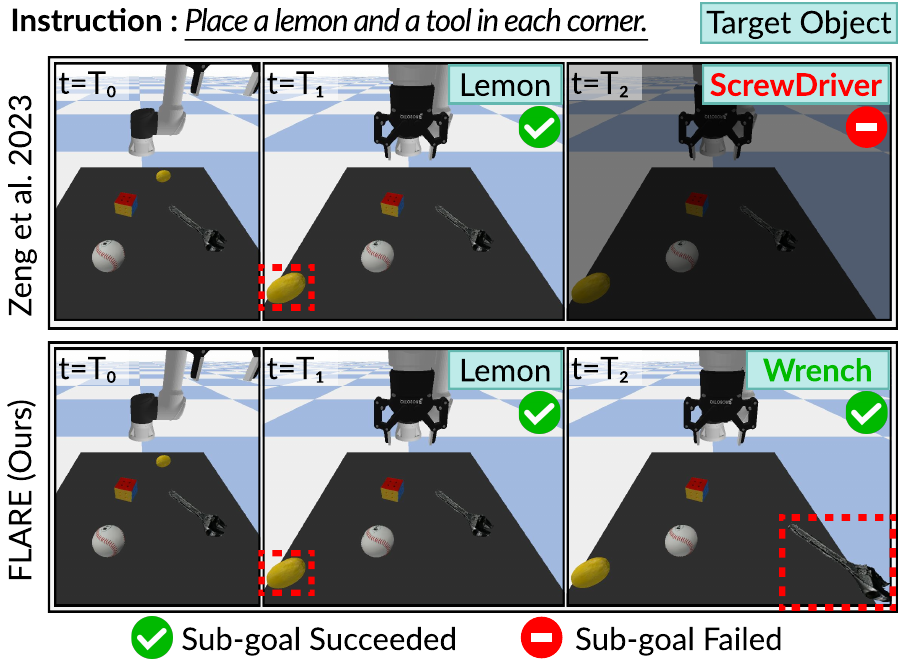}
    \caption{
        \textbf{An example of robotic task applications.}
        Baseline model~\cite{zeng2022socratic} generates an ungrounded plan due to ambiguous instructions (\eg, \textit{tool}).
        In contrast, \method generates a grounded plan and successfully exectues actions.
        }
    \vspace{-0.7em}
    \label{fig:robot_quali}
\end{figure}

\subsection{Application in Robotic Task Planning}
\label{sec:additional_robot}
We demonstrate the generalizability of the proposed \method to other robotic task applications.
Specifically, we use a simulated tabletop environment with an UR5 robot arm and illustrate a comparison between \method and the baseline model in Figure~\ref{fig:robot_quali}.
We choose \cite{zeng2022socratic} as the baseline model for its effectiveness in few-shot robot planning.
Both models use GPT-3.5 as an LLM to generate subgoals and employ a privileged low-level policy which uses the environment's metadata for end effector pose prediction.

We observe that \method successfully rearranges objects as instructed, demonstrating its capability in planning for grounded execution.
In contrast, the baseline~\cite{zeng2022socratic} fails due to an ungrounded plan (\eg, attempting to pick a \textit{ScrewDriver} that is not present in the environment).


\section{Conclusion}

We propose \method with a multi-modal planner that reflects both environmental status by visual input and language instruction to generate detailed plans (\ie, subgoals) to accomplish a long-horizon tasks with a few data.
Additionally, it revises only the subset of the subgoals that are incorrect to generate physically grounded plans without using LLMs, leading to computationally efficient replanning.
We empirically validate the effectiveness of the proposed components in ALFRED~\cite{shridhar2020alfred} and observe that our \method outperforms the state-of-the-art methods in few-shot settings by significant margins in all metrics.

\vspace{-0.2em}
\paragraph{Limitations and future work.}
Although our method requires a very few fraction of training data (0.5\%), it still requires the training data.
We aim to develop an agent that learns about environments through exploration, assisted by large language models, without needing any training data.

\section*{Acknowledgments}
This work was partly supported by the IITP grants (No.RS-2022-II220077, No.RS-2022-II220113, No.RS-2022-II220959, No.RS-2022-II220871, No.RS-2021-II211343 (SNU AI), No.RS-2021-II212068 (AI Innov. Hub), No.RS-2022-II220951) funded by the Korea government(MSIT).

\bibliography{aaai25}

\clearpage
\appendix

\section*{Supplemenatry Materials for \\ Multi-Modal Grounded Planning and Efficient Replanning
For Learning Embodied Agents with A Few Examples}
~
~
~
\newcommand{\origref}[1]{[{#1}]}


\section{Details of ALFRED Benchmark}  
\label{sup:benchmark_detail}

The ALFRED (Action Learning From Realistic Environments and Directives) benchmark~(Shridhar et al. 2020) is a benchmark designed to test embodied agents in understanding and executing a variety of natural language instructions within a simulated household environment.
It consists of seven task types with $115$ distinct object types.
The aim is to understand natural language instructions and to complete long-horizon tasks.
To satisfy predefined task conditions, an agent must execute a sequence of actions and generate object masks for interacting with objects in the environment.
Each task comes with a high-level natural language instruction, accompanied by detailed, low-level directives that specifically guide the agent's actions.
Failure to meet any of these conditions results in the task being deemed unsuccessful.

ALFRED provides three distinct splits: `training,' `validation,' and `test.'
Agents can be trained with the `training' split and can have their approaches verified within the `validation' split where they have access to the ground-truth information of the tasks in those splits.
The agents are then evaluated in the `validation' and `test' splits, without any ground-truth data pertaining to the tasks.

At every timestep, an agent within the environment operates based on an egocentric RGB visual input in the shape of the $300\times300$ image.
From this input, the agent must select an appropriate action from a predefined action space, which includes both navigational and object interaction commands.
Alongside these actions, the agent also generates a binary object mask to specify interaction targets, corresponding to the same resolution (\ie, $300\times300$) as the visual input.

Action commands consist of navigational actions such as \textsc{MoveAhead}, \textsc{RotateRight}, \textsc{RotateLeft}, \textsc{LookUp}, and \textsc{LookDown}. Interaction actions encompass \textsc{PickupObject}, \textsc{PutObject}, \textsc{OpenObject}, \textsc{CloseObject}, \textsc{ToggleObjectOn}, \textsc{ToggleObjectOff}, and \textsc{SliceObject}.
The action \textsc{Stop} signifies the agent's decision to end the task, ideally once all conditions are met.

ALFRED employs multiple metrics to comprehensively quantify an agent's performance.
The primary metric is the Success Rate (SR), which measures the proportion of fully completed tasks.
A secondary metric, the Goal-Condition Success Rate (GC), accounts for partially completed tasks where the agent satisfies some but not all of the required conditions.
Finally, Path Length Weighted (PLW) scores adjust the SR and GC metrics (\ie, PLWSR and PLWGC) based on the length of the action sequences undertaken by the agent.
Expert demonstrations, which use the shortest path for navigation without unnecessary exploration, are generally considered optimal.
If an agent takes twice as long as the expert to complete a task, it gets only half the credit.

\section{Additional Implementation Details}
\subsection{Semantic Mapping } 
\label{sub:semanticmap}
We first predict an instance segmentation and a depth from the agent's egocentric RGB input.
Then we transform these predictions into a point cloud, where each point is assigned a corresponding semantic label, resulting in labeled voxels.
To generate the 2D semantic map, we finally aggregate these 3D voxels by summing them across their vertical dimension
At every step, the model continuously updates the global map by incorporating the newly obtained partial maps.

\subsection{LLM and Prompt}
We use four large language models for the implementation of \method.
For proprietary models\footnote{https://platform.openai.com/docs/models}, we use \textsc{gpt-3.5-turbo-instruct} (referred to as GPT-3.5 in the main text) and \textsc{gpt-4-0125-preview} (referred to as GPT-4 in the main text).
We set the temperature to $0$ to ensure reproductivity and apply a logit bias of $0.1$ to all allowable output tokens (\eg, allowable A, O, R in Sec.~\ref{sec:subgoal_triplet}).
For open-source models, we use \textsc{llama-2-13b-chat}\footnote{https://huggingface.co/meta-llama/Llama-2-13b-chat} (referred to as LLaMa2-13B in the main text) and \textsc{vicuna-13b-v1.5}\footnote{https://huggingface.co/lmsys/vicuna-13b-v1.5} (referred to as Vicuna-13B in the main text).
For both open-source models, we set the temperature value to default.

\begin{figure}[t!]
    \centering
    \includegraphics[width=0.98\linewidth]{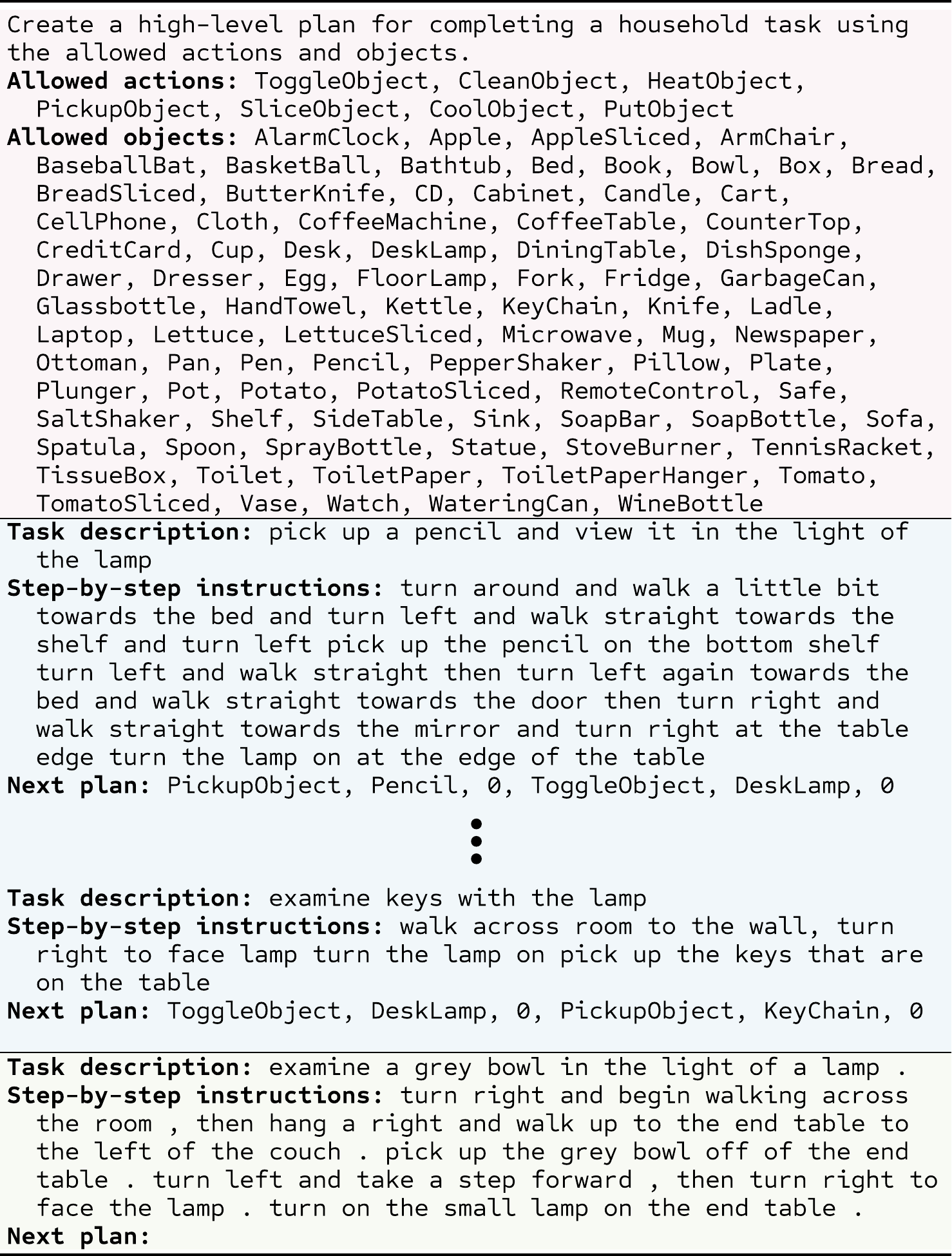}
    \caption{\textbf{An example of prompt used in Multi-Modal Planner.}
    A block denoted in pink provides an explanation of the task.
    A block denoted in blue serves as an in-context example.
    Finally, a block denoted in green represents the current task.
    }
    \label{fig:prompt_ex}
    \vspace{-1.5em}
\end{figure}

We provide an example of prompt used in MMP (Sec.~\ref{sec:mmp}) in Figure~\ref{fig:prompt_ex}.
We first provide an explanation of the task and a list of all allowed actions and objects (block denoted with pink).
Then, we present retrieved examples as in-context examples with headers `Task description,' `Step-by-step instructions,' and `Next plan' (block denoted with blue).
Finally, we show the current task in the same format as in-context examples, leaving the blank after the `Next plan' header (block denoted with green).
For `Goal instruction only' setup, which prohibits the use of step-by-step instructions, we remove step-by-step instructions in the prompt for both retrieved example and the current task.

\begin{figure}[t!]
    \centering
    \includegraphics[width=1\linewidth]{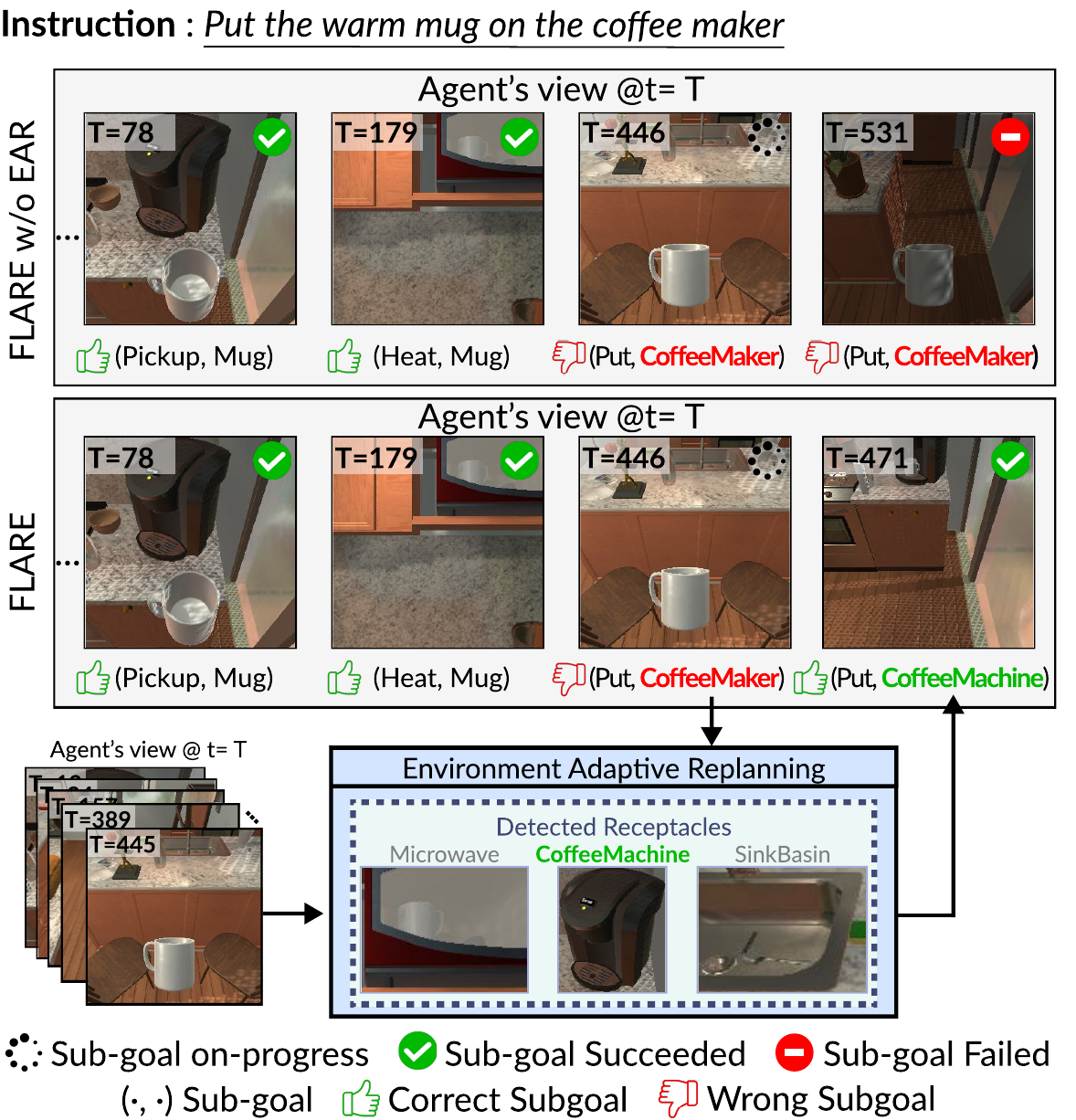}
    \caption{
        \textbf{A qualitative example of our agent with and without `Environmental Adaptive Replanning' (EAR).}
        \method w/o EAR fails to locate \textit{CoffeMaker}, leading to task failure.
        In contrast, \method requests replanning to EAR and it replaces partially incorrect subset of the subgoal (\ie, \textit{CoffeeMaker} $\rightarrow$ \textit{CoffeeMachine}).
    }
    \label{fig:supple_quali1}
    \vspace{-1.5em}
\end{figure}

\section{Additional Qualitative Examples} 
\label{sup:add_quali}
\begin{figure}[t]
    \centering
    \includegraphics[width=.98\linewidth]{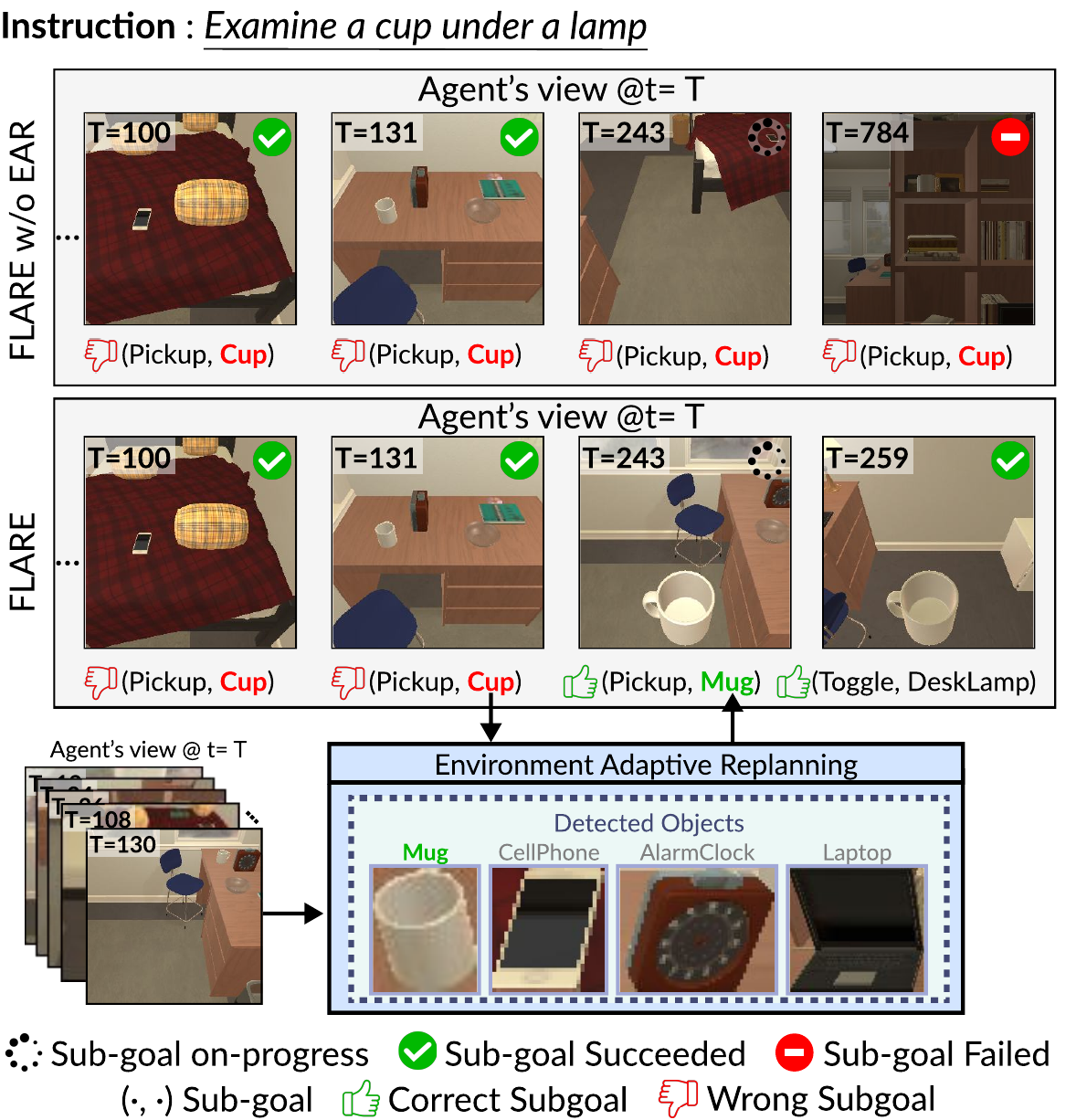}
    \caption{
        \textbf{Another qualitative example of our agent with and without `Environmental Adaptive Replanning' (EAR).}
        \method w/o EAR fails to find a \textit{Cup} and achieves zero goal conditions (GC).
        Conversely, \method requests replanning to EAR and it replaces a partially incorrect subset of the subgoal (\ie, \textit{Cup} $\rightarrow$ \textit{Mug}).
    }
    \label{fig:supple_quali2}

\end{figure}

We provide additional qualitative examples in Figures~\ref{fig:supple_quali1} and~\ref{fig:supple_quali2} in the same manner as in Figure~\ref{fig:corr_quali}.
To successfully complete a task, not even a single subgoal should fail.
Figure~\ref{fig:supple_quali1} shows an example of an agent without EAR (\ie, \method w/o EAR) successfully finding the mug and heating it up.
However, it fails to place the object in the intended receptacle and the task is considered a failure.
In contrast, as \method cannot locate \textit{CoffeeMaker}, it requests replanning to EAR.
EAR infers that \textit{CoffeMachine} is the most semantically similar to \textit{CoffeeMaker}, and revises its subgoal.

If the first subgoal is incorrect, an agent would fail to achieve subsequent subgoals, aimlessly searching for an undetectable object.
Figure~\ref{fig:supple_quali2} illustrates that the agent without EAR (\ie, \method w/o EAR) achieves zero goal conditions (GC).
Due to the operator misidentifying the object as a \textit{Cup}, not a \textit{Mug}, based on its similar appearances and functionality, the agent roams the room in endless search of a \textit{Cup}.

In contrast, an agent equipped with EAR (\ie, \method) noticing that the instructed \textit{Cup} is misleading, replaces it with a \textit{Mug} that is presented in the scene.
After adapting the plan accordingly, the agent proceeds to perform the remaining subgoals and ultimately completes the task.

\section{Robotic Application} 
\label{sup:additional_robot}
We evaluate the effectiveness of \method in a simulated robotic task using PyBullet\footnote{https://pybullet.org/} as a simulation environment and an UR5 robot equipped with a parallel gripper.
We first create a small training dataset with $65$ instruction and subgoal pairs using GPT-4o.
In each scenario, between $3$ to $6$ objects were randomly placed on a tabletop, providing varying tasks for the robot.
The goal of the UR5 robot is to follow the instructions, such as ``Place a lemon and a tool in each corner,'' by accurately picking and placing objects accurately.
Both \method and the baseline model~(Zeng et al. 2023) use \textsc{gpt-3.5-turbo-instruct} (referred to as GPT-3.5 in the main text) as an LLM to generate high-level subgoals.
For low-level action prediction, both models employ a privileged policy.
Given each high-level subgoal, the robot uses the simulated environment's metadata to obtain target coordinates, actuating its joints via inverse kinematics.


\null
\vfill
\end{document}